\documentclass[letterpaper, 10 pt, conference]{ieeeconf}  
\usepackage{hyperref}
\usepackage{graphicx}
\usepackage{amsmath,amssymb} 
\usepackage{color}
\usepackage{cleveref}
\usepackage{array}
\usepackage{pgfplots}
\pgfplotsset{compat=1.16}
\usepackage{pgfplotstable}
\usepackage{subcaption}

\usepackage{multirow}
\usepackage{booktabs}
\usepackage{threeparttable}
\usepackage{pifont} 
\usepackage{amssymb}

\IEEEoverridecommandlockouts                              

\title{\LARGE \bf
PFSD: A Multi-Modal Pedestrian-Focus Scene Dataset for Rich Tasks in Semi-Structured Environments
}

\author{Yueting Liu$^{1,2}$, Hanshi Wang$^{2,3}$, Zhengjun Zha$^{1}$, Weiming Hu$^{2,3}$ and Jin Gao$^{2,3*}$
\thanks{$^{1}$School of Information Science and Technology, University of Science and Technology of China, Anhui, China
        {\tt\small lytmatters @mail.ustc.edu.cn zhazj@ustc.edu.cn}}%
\thanks{$^{2}$State Key Laboratory of Multimodal Artificial Intelligence Systems (MAIS), CASIA, Beijing, China
        {\tt\small hanshi.wang.cv@outlook.com, 
        \{wmhu, jin.gao\} @nlpr.ia.ac.cn}}%
\thanks{$^{3}$School of Artificial Intelligence, University of Chinese Academy of Sciences, Beijing, China}
\thanks{$^{*}$Corresponding author.}%
}

\begin{document}

\maketitle
\thispagestyle{empty}
\pagestyle{empty}

\begin{abstract}

Recent advancements in autonomous driving perception have revealed exceptional capabilities within structured environments dominated by vehicular traffic. However, current perception models exhibit significant limitations in semi-structured environments, where dynamic pedestrians with more diverse irregular movement and occlusion prevail. We attribute this shortcoming to the scarcity of high-quality datasets in semi-structured scenes, particularly concerning pedestrian perception and prediction.
In this work, we present the multi-modal Pedestrian-Focused Scene Dataset(PFSD), rigorously annotated in semi-structured scenes with the format of nuScenes. PFSD provides comprehensive multi-modal data annotations with point cloud segmentation, detection, and object IDs for tracking. It encompasses over 130,000 pedestrian instances captured across various scenarios with varying densities, movement patterns, and occlusions.
Furthermore, to demonstrate the importance of addressing the challenges posed by more diverse and complex semi-structured environments, we propose a novel Hybrid Multi-Scale Fusion Network (HMFN). Specifically, to detect pedestrians in densely populated and occluded scenarios, our method effectively captures and fuses multi-scale features using a meticulously designed hybrid framework that integrates sparse and vanilla convolutions. Extensive experiments on PFSD demonstrate that HMFN attains improvement in mean Average Precision (mAP) over existing methods, thereby underscoring its efficacy in addressing the challenges of 3D pedestrian detection in complex semi-structured environments. Coding and benchmark are available at: \nolinkurl{https://github.com/VSlabRobotics/PFSD.git}.

\end{abstract}

\section{INTRODUCTION}

In recent years, autonomous driving has advanced rapidly, driven by increasingly sophisticated perception algorithms. High-quality datasets serve as the cornerstone of perception algorithms, enabling models to learn intricate details, adapt to dynamic conditions, and ultimately achieve robust performance by offering the necessary complexity and diversity of environments and objects. As state-of-the-art algorithms approach improvement on widely used benchmarks, there is a growing imperative for datasets that more accurately reflect the complexity and unpredictability of real-world scenarios. This naturally raises the question: \textit{What type of dataset can further catalyze the evolution of autonomous driving?}
\begin{figure}
\centering
\includegraphics[width=8.5cm, height=4.5cm]{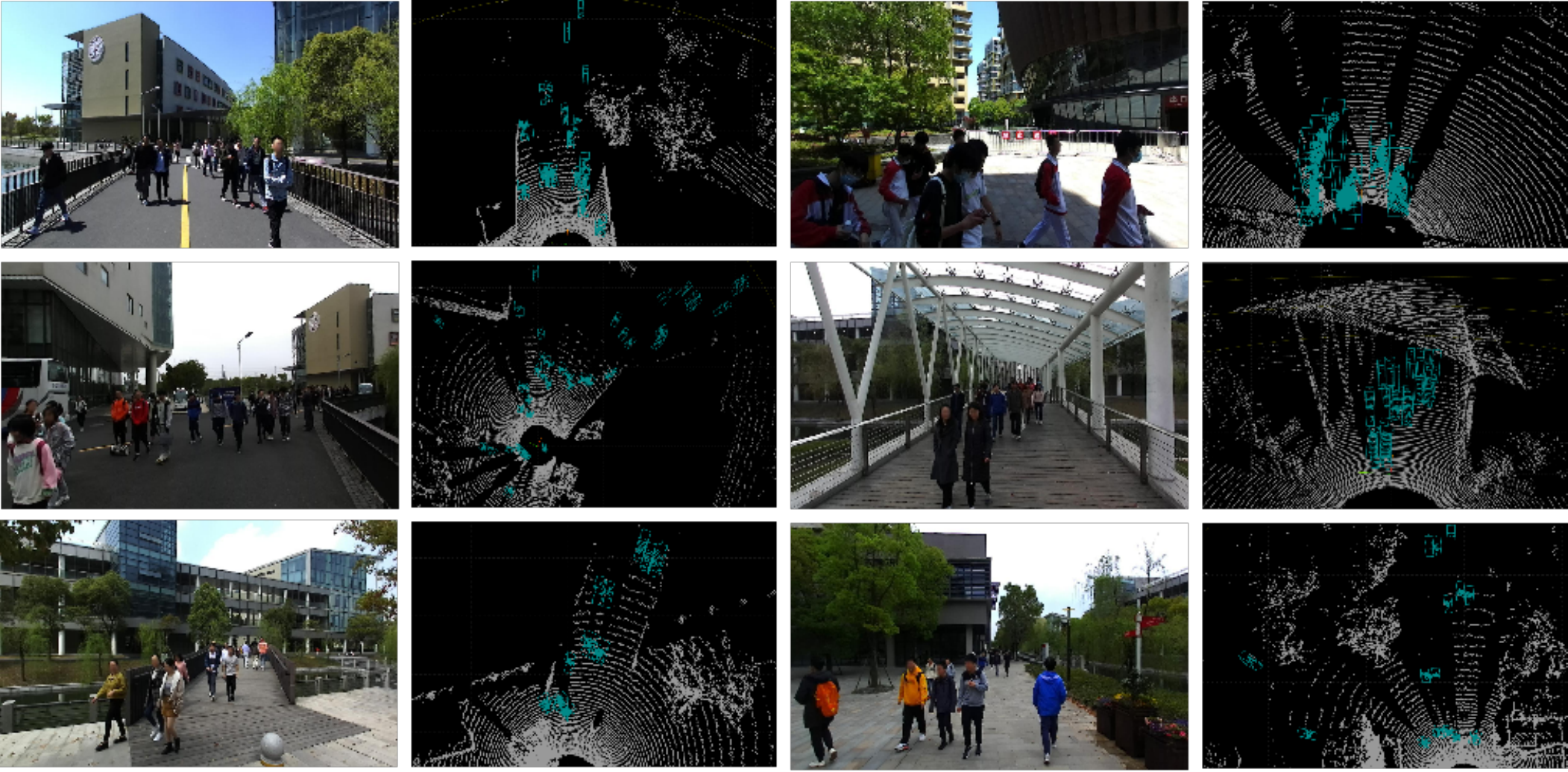}
\caption{Semi-structured scene examples of PFSD with different densities provide extensive annotations for pedestrians across a wide range of scenes and environmental conditions.}
\label{fig:densities}
\end{figure}

On the one hand, recent developments in embodied intelligence underscore the importance of training agents in realistic, less controlled scenarios. Such environments demand that autonomous systems frequently interact with pedestrians. A re-examination of existing datasets reveals that they can be roughly grouped into structured or unstructured categories. Structured datasets (\textit{e.g.}, KITTI~\cite{kitti}, ApolloScape~\cite{huang2018apolloscape}, Waymo~\cite{waymo}, and others~\cite{h3d,lyftl5,chang2019argoverse,A3D,a2d2,nuscenes,cirrus,pandaset,once}) typically focus on well-defined, rule-governed roadways, emphasizing vehicular traffic. However, these datasets often oversimplify pedestrian perception and prediction tasks by only including pedestrians in relatively straightforward contexts with regular motion. This restricts the applicability of models trained on such data in more complex, pedestrian-centric scenes. Conversely, unstructured datasets (\textit{e.g.}, RELLIS-3D~\cite{rellis}) present irregular backgrounds and complex distributions of objects. While these datasets broaden the range of encountered scenarios, their scarcity of pedestrian data diminishes their practical value for training autonomous driving models.

\begin{table*}[h!]
\centering
\caption{\small Comparison of multi-modal Pedestrian-Focus Scene Dataset (PFSD) with existing datasets. Pedes-num stands for the total number of pedestrian identification data in each dataset. Pedes/Fr stands for the average number of pedestrian identifiers per point cloud frame. Density-2/5/10 stands for the average number of pedestrians of PFSD within radii of 2m, 5m, and 10m around each individual.}
\label{tab:datasets}
\resizebox{\textwidth}{!}{%
\begin{threeparttable}
\begin{tabular}{@{}llcccccccc@{}}
\toprule
\textbf{Dataset} & \textbf{Scene Type}       & \textbf{LiDAR scans} & \textbf{Beam} & \textbf{Pedes-num}  & \textbf{Pedes/Fr} & \textbf{Density-2/5/10} & \textbf{Segmentation} & \textbf{3D detection} & \textbf{Tracking} \\ \midrule
KITTI            & Structured                & 15k                  & 64            & 4.5k                & 0.3                    & 0.5/1.3/2.3             & $\times$\tnote{§}                 & $\checkmark$            & $\checkmark$        \\
ApolloScape\tnote{*} & Structured                & 12k / 5.6k          & -             & 60.1k / 30.4k       & 5                      & 0.3/0.6/1.1            & $\checkmark$            & $\checkmark$            & $\checkmark$        \\
H3D              & Mainly Structured                & 27k                  & 64            & 280k                & 10                     & 1.5/4.0/7.2            & $\times$                & $\checkmark$            & $\checkmark$        \\
Lyft L5\tnote{†}  & Structured                & 46k                  & 40            & 210k                & 4.6                    & -                      & $\checkmark$            & $\checkmark$            & $\checkmark$        \\
Argoverse        & Structured                & 22k                  & 32            & 110k                & 5                      & -                      & $\times$                & $\checkmark$            & $\checkmark$        \\
A*3D\tnote{†}     & Structured                & 39k                  & 64            & 20k                 & 0.5                    & -                      & $\times$                & $\checkmark$            & $\times$            \\
A2D2\tnote{‡}     & Structured                & 12.5k                & 16            & 4.3k                & 0.4                    & 0.3/0.5/0.7            & $\checkmark$            & $\checkmark$            & $\times$            \\
nuScenes         & Structured                & 40k                  & 32            & 208k                & 5                      & 0.7/1.6/2.7            & $\times$\tnote{§}       & $\checkmark$            & $\checkmark$        \\
Waymo            & Structured                & 230k                 & 64            & 2.8M                & 12                     & 1.0/2.9/5.6            & $\times$\tnote{§}       & $\checkmark$            & $\checkmark$        \\
Cirrus           & Structured                & 6.2k                 & 64            & 11.8k               & 1.9                    & 0.4/0.8/1.2            & $\times$                & $\checkmark$            & $\times$            \\
PandaSet\tnote{†} & Mainly Structured                & 8.2k                 & 64            & -                   & 30.2                   & -                      & $\checkmark$            & $\checkmark$            & $\times$            \\
ONCE\tnote{*}     & Structured                & 16k / 8.2k          & 40            & - / 37k             & 4.5                    & 0.4/0.8/1.2            & $\times$                & $\checkmark$            & $\times$            \\
RELLIS-3D      & Unstructured           & 13k                   & 64           & -                & -                     & -           & $\checkmark$            & $\times$             & $\times$         \\
PFSD (ours)      & Semi-Structured           & 4k                   & 128           & 133k                & 32                     & 2.6/6.0/11.6           & $\checkmark$            & $\checkmark$            & $\checkmark$        \\ 
\bottomrule
\end{tabular}
\begin{tablenotes}
\footnotesize
\item[*] No test set available. LiDAR scans and pedestrian numbers are split into total and annotated data.
\item[†] Dataset currently unavailable for download.
\item[‡] Subsets of annotated data used for statistics.
\item[§] Extended versions available with segmentation capabilities.
\end{tablenotes}
\end{threeparttable}%
} 
\end{table*}

This gap suggests the necessity of an alternative: datasets that lie between these two extremes. Hence, we propose assembling datasets from semi-structured scenarios, recorded in environments with partial but not fully consistent road and traffic guidance, and ensuring that pedestrian annotations remain central. Specifically, based on the collected data in pedestrian-focused semi-structured scenes in the work of~\cite{stcrowd}, we further rigorously annotated it with point cloud segmentation, detection, and object IDs for tracking with the format of nuScenes to facilitate various camera-based, lidar-based or multi-modal fusion-based  3D perception tasks. Our new multi-modal pedestrian-focused scenes dataset (PFSD) can bridge the gap between highly structured driving environments and unstructured, open-ended scenarios. These scenes shown in \Cref{fig:densities} combine the regularity of structured environments with the dynamism of unstructured environments. 
Another key feature of PFSD is its high-density pedestrian annotations in semi-structured scenes where pedestrians exhibit dynamic, unpredictable, and complex behavior, which is central to understanding real-world interactions. 
Meanwhile, PFSD supports tasks such as detection, tracking, and segmentation simultaneously, which allows researchers to develop unified models that can handle segmentation, detection, and tracking at the same time.

PFSD provides diverse and challenging semi-structured scenes for object detection, tracking, and segmentation.
For object detection, PFSD enables models to handle background variations and dynamic scenes. Unlike structured datasets like KITTI~\cite{kitti} with fixed scene patterns that limit generalization, PFSD’s high-density pedestrian annotations emphasize pedestrian movement in complex urban settings.
As for object tracking, PFSD captures dynamic object interactions which are often limited in structured datasets constrained by traffic rules. It presents complex tracking scenarios with pedestrian interaction, occlusion, and overlap, providing long-term data to address track breakage and target loss in dense pedestrian scenes, thereby enhancing multi-object tracking performance. Moreover, PFSD employs fixed-point static data acquisition to ensure stable and long-term tracking.
In the segmentation task, PFSD helps models capture details, overcoming the limitations of structured datasets in real-world applications. It provides diverse training environments where pedestrians interact with both static structures and dynamic elements to improve model performance.

On the other hand, the release of large-scale and high-quality datasets has significantly advanced AI research, particularly in autonomous driving, providing essential training resources and driving algorithmic improvements. In 3D object detection and tracking, the nuScenes dataset has become a key resource due to rich sensor data, detailed annotations, and comprehensive metadata. Despite its strengths, nuScenes also lacks coverage in dense scenes.

PFSD with the format of nuScenes also addresses this limitation. While the original dataset~\cite{stcrowd} captures dense pedestrian movement and interactions, it lacks structured metadata and systematic organization. Meanwhile, it does not provide segmentation data and is incompatible with any standardized data format. These not only restrict data querying capabilities but also hinder integration with advanced frameworks such as OpenPCDet and MMDetection.

In addition to organizing detection and tracking data to the nuScenes standard, building upon~\cite{stcrowd}, we further incorporated point cloud segmentation data, addressing limitations of the original dataset. To the best of our knowledge, this is the first dataset introducing dense pedestrian scenes into the nuScenes framework. By leveraging segmentation data, we not only expand the applicability of~\cite{stcrowd}, enabling segmentation, detection, and tracking tasks to be evaluated simultaneously in semi-structured dense scenarios, but also enhance the precision of 3D bounding boxes based on segmentation results. This improvement may significantly contribute to the refinement of detection, trajectory tracking, and prediction research. This transformation also ensures compatibility with existing tools, facilitating algorithm development for dense scenes. The standardized format enables cooperation with frameworks like OpenPCDet and MMDetection and supports object detection models as CenterPoint~\cite{centerpoint}, TransFusion~\cite{transfusion}, Bevformer~\cite{bevformer}, PETR~\cite{petr}, as well as tracking methods like AB3DMOT~\cite{ab3dmot}, QTrack~\cite{qtrack}, and SimpleTrack~\cite{simpletrack}, which all related to nuScenes. This expansion not only improves model validation in dense scenarios but also broadens the dataset’s applicability, laying the foundation for intelligent perception and automated decision-making in urban dense scenes.

Additionally, in semi-structured scenes with high pedestrian density, such as PFSD, fine-grained features need to be extracted to overcome the challenges posed by occlusion in dense environments. Traditional single-scale feature extraction methods including pillarNet~\cite{shi2022pillarnet}
 often fail to capture detailed pedestrian information, especially in crowded and complex semi-structured scenes. To address these challenges, we propose the Hybrid Multi-Scale Fusion Network (HMFN), a novel architecture specifically designed to leverage the rich pedestrian data from multi-scale features. HMFN can act as a simple baseline that combines features extracted by PillarNet~\cite{shi2022pillarnet} across different scales through advanced upsampling and attention mechanisms before passing them to a CenterHead detection module~\cite{centerpoint}. This design enables precise pedestrian detection by combining fine-grained and macro-level features, even in challenging scenarios. Experiments show HMFN significantly improves mean Average Precision (mAP) compared to existing methods, proving its effectiveness in 3D pedestrian perception. Similarly, this intuitive idea of extracting multi granularity features can also be applied to segmentation and tracking tasks.

Our contributions are summarized as follows:

    - We introduce the multi-modal Pedestrian-Focused Scene Dataset (PFSD), first introducing cross-modal dense crowd scenes within the nuScenes format data and simultaneously supporting detection, tracking, and segmentation tasks.
    
    - PFSD balances structured predictability with unstructured complexity. It provides dense pedestrian annotations, enhancing coverage of high-density pedestrian environments while ensuring compatibility with existing benchmarks.
    
    - To get rich information in semi-structured scenes, we propose the Hybrid Multi-Scale Fusion Network (HMFN) to significantly enhance the detection of fine-grained pedestrian features in complex and crowded environments.

\section{RELATED WORK}

\subsection{Multi-modal datasets}
The availability of high-quality datasets has significantly advanced research in 3D perception tasks, such as segmentation, object detection, and tracking. However, existing datasets often differ in their design focus, environmental settings, and annotation granularity. To provide a structured overview, we categorize related datasets into three types.

\textbf{Structrured.} Structured datasets, designed for controlled environments like urban roads, focus on vehicular scenarios with annotations for vehicles and traffic-related objects. The KITTI dataset~\cite{kitti} pioneered 3D object detection with LiDAR and image data, but the limited size and pedestrian diversity restrict its use in studying pedestrian dynamics. Datasets like nuScenes~\cite{nuscenes} and Waymo Open Dataset~\cite{waymo} expanded the scope with multimodal data, including LiDAR, cameras, and radar, alongside 3D annotations. Other structured datasets, such as ApolloScape~\cite{huang2018apolloscape} and Argoverse~\cite{chang2019argoverse}, offer traffic-related annotations but lack focus on complex pedestrian dynamics. However, these datasets are primarily vehicle-centric, with limited attention to pedestrian behavior in dense or unstructured settings.

\textbf{Semi-Structured.} Semi-structured datasets capture environments like campus paths and parking lots, where irregular pedestrian flows and occlusions present unique challenges. Datasets such as H3D~\cite{h3d} and PandaSet~\cite{pandaset} offer multimodal annotations for 3D detection and segmentation in semi-structured settings but are not abundant. Our dataset, PFSD, fills this gap by focusing on dynamic pedestrian interactions in irregular environments, with rich annotations for segmentation, detection, and tracking.

\textbf{Unstructured.} Unstructured environments feature minimal organization and no clearly defined pathways, such as forests, disaster zones, and rural areas. Rellis-3D~\cite{rellis}, which focuses on segmentation and obstacle detection in unstructured rural settings but typically lacks pedestrian annotations and fails to capture pedestrian-specific dynamics. This makes it unsuitable for studying pedestrian perception in mixed environments, leaving a gap in research that datasets like PFSD aim to address.

\begin{figure}[t!]
    \centering
    \includegraphics[width=8.5cm]{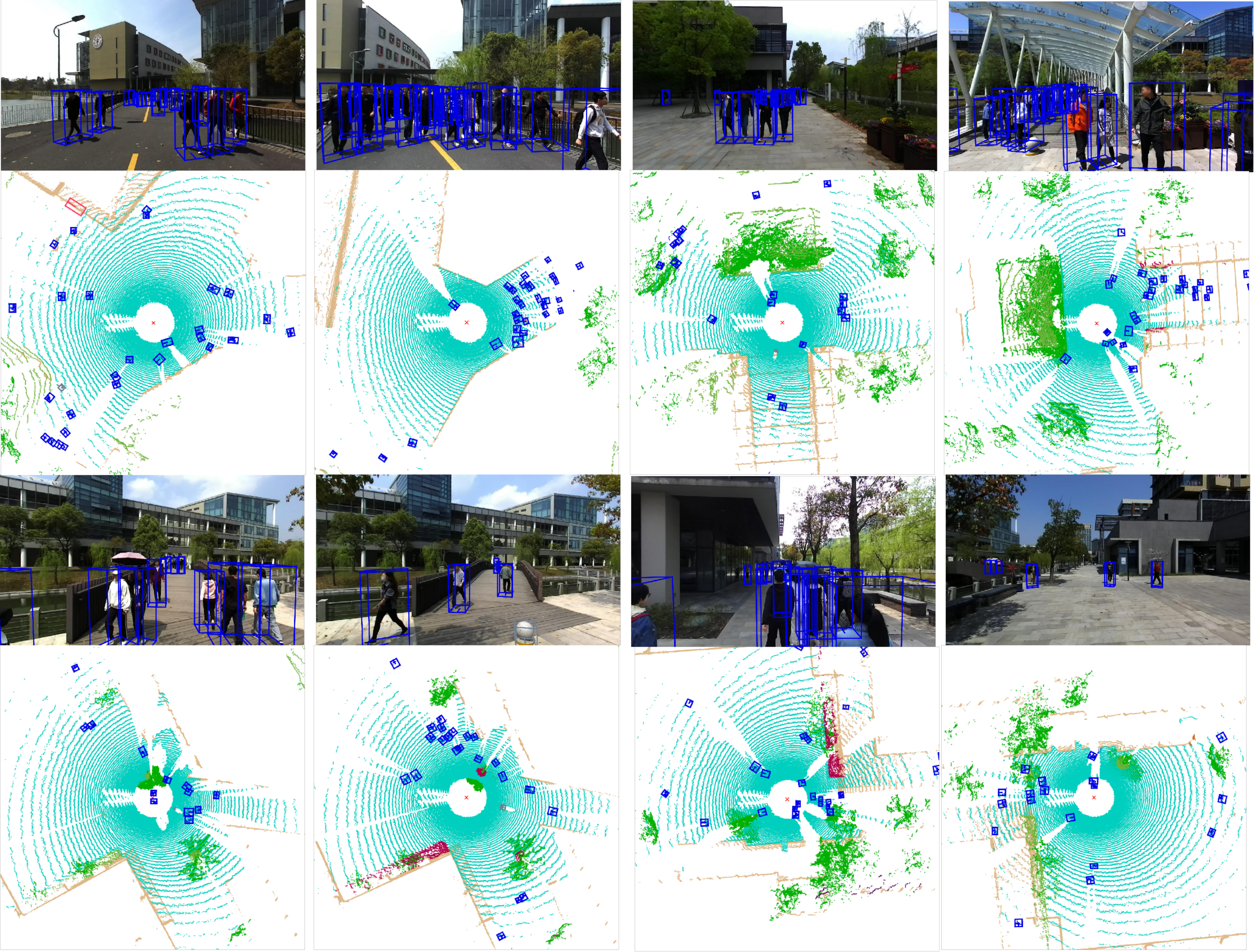}
    \caption{Illustration of pedestrian detection in various semi-structured scenes, highlighting the richness of pedestrian annotations and the complexity of semi-structured environments in both images and point clouds.}
    \label{fig:detection}
\end{figure}

\subsection{3D object detection}

3D object detection has made significant strides in recent years and various approaches have been developed to handle the unique challenges of point-cloud data, which are inherently unstructured and sparse. The primary approaches can be categorized into different methods, each offering distinct advantages and facing specific limitations. 

\textbf{Point-based} methods operate directly on raw point clouds, preserving fine details. PointNet~\cite{qi2017pointnet} introduced this approach using MLPs but struggled with local structure. PointNet++~\cite{qi2017pointnet++} improved this by introducing multi-scale grouping. PointRCNN~\cite{shi2019pointrcnn} further refined this with a two-stage 3D bounding box prediction to enhance accuracy.

\textbf{Voxel-based} methods convert point clouds into voxel grids and use 3D convolutions. VoxelNet~\cite{zhou2018voxelnet} showed the potential of 3D CNNs, though it was computationally expensive. SECOND~\cite{yan2018second} reduced computational costs using sparse convolutions, enabling real-time tasks.

\textbf{Pillar-based} methods convert point clouds into vertical pillars for efficient 2D convolutions, improving detection speed. PointPillars~\cite{lang2019pointpillars} pioneered this approach, while PillarNet~\cite{qi2017pointnet} enhanced it with sparse convolutions. 

\section{Benchmark Detail}

\subsection{Standardized NuScenes Data Organization}
To enhance usability, we organized the dataset~\cite{stcrowd} into the nuScenes format, addressing its initial limitations in data organization and accessibility. Unlike the original dataset, which stored scene and frame data in a simple list format, our structured conversion ensures consistency, scalability, and seamless interoperability with established frameworks.

First, a well-structured format enables efficient data management and retrieval. By following the nuScenes schema, PFSD supports timestamped multi-frame organization and precise spatial-temporal alignment, making it significantly easier to conduct multi-frame reasoning for tasks like long-term tracking and trajectory prediction. 
Second, standardization fosters compatibility with existing autonomous driving and perception frameworks. PFSD can be directly integrated into widely used pipelines such as OpenPCDet and MMDetection, allowing researchers to experiment with state-of-the-art 3D object detection and multi-object tracking, enabling robust evaluation in high-density pedestrian scenes.  
Furthermore, structuring PFSD with nuScenes format facilitates model validation under real-world urban challenges. Traditional structured datasets focus on vehicle-centric scenes with well-defined road structures, whereas PFSD introduces dense pedestrian interactions, heavy occlusion, and dynamic scene variations, making it a valuable benchmark for testing perception algorithms under crowded conditions.

\subsection{Object Detection in Semi-Structured Environments}
In object detection tasks, when scenes in an object detection dataset exhibit diversity, complexity, and realism, while covering various target features and background variations in different environments, it helps improve the model's generalization ability to diverse scenes. 

Structured datasets, such as KITTI~\cite{kitti}, ApolloScape~\cite{huang2018apolloscape}, and Waymo~\cite{waymo}, focus on autonomous driving with fixed object layouts and relatively stable scenes. While effective in controlled settings, they lack the diversity needed for real-world generalization. In contrast, unstructured datasets like RELLIS-3D~\cite{rellis} feature cluttered backgrounds and complex object relationships but often introduce noise and ambiguity, posing challenges for stable model performance.

\begin{figure}[t!]
    \centering
    \includegraphics[width=8.5cm, height=4.5cm]{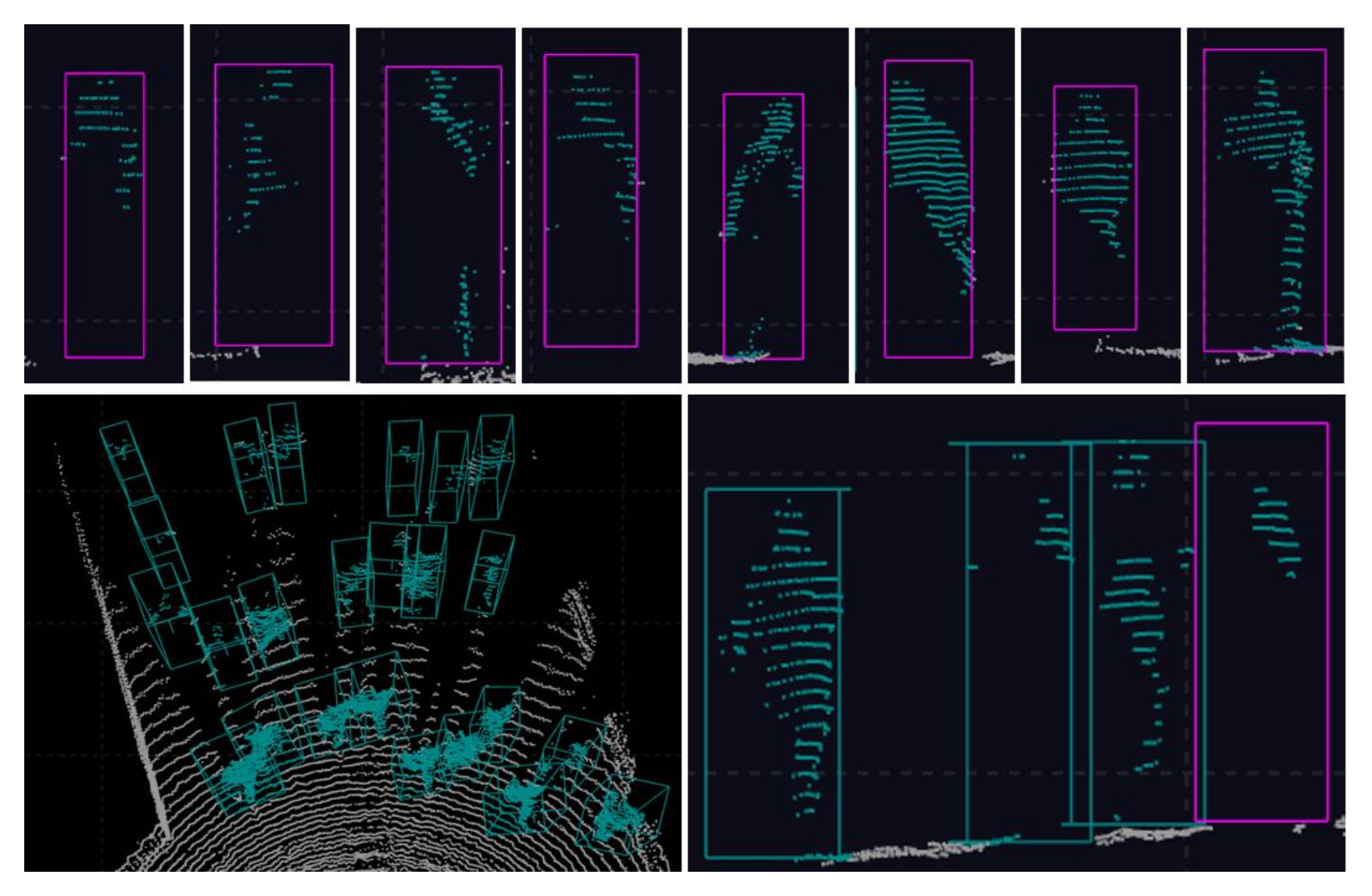}
    \caption{Visualization of varying degrees of pedestrian occlusion in crowded semi-structured environments, demonstrating the impact of occlusion. }
    \label{fig:occlusion}
\end{figure}

\begin{figure}[t!]
    \centering
    \includegraphics[width=8.5cm, height=6cm]{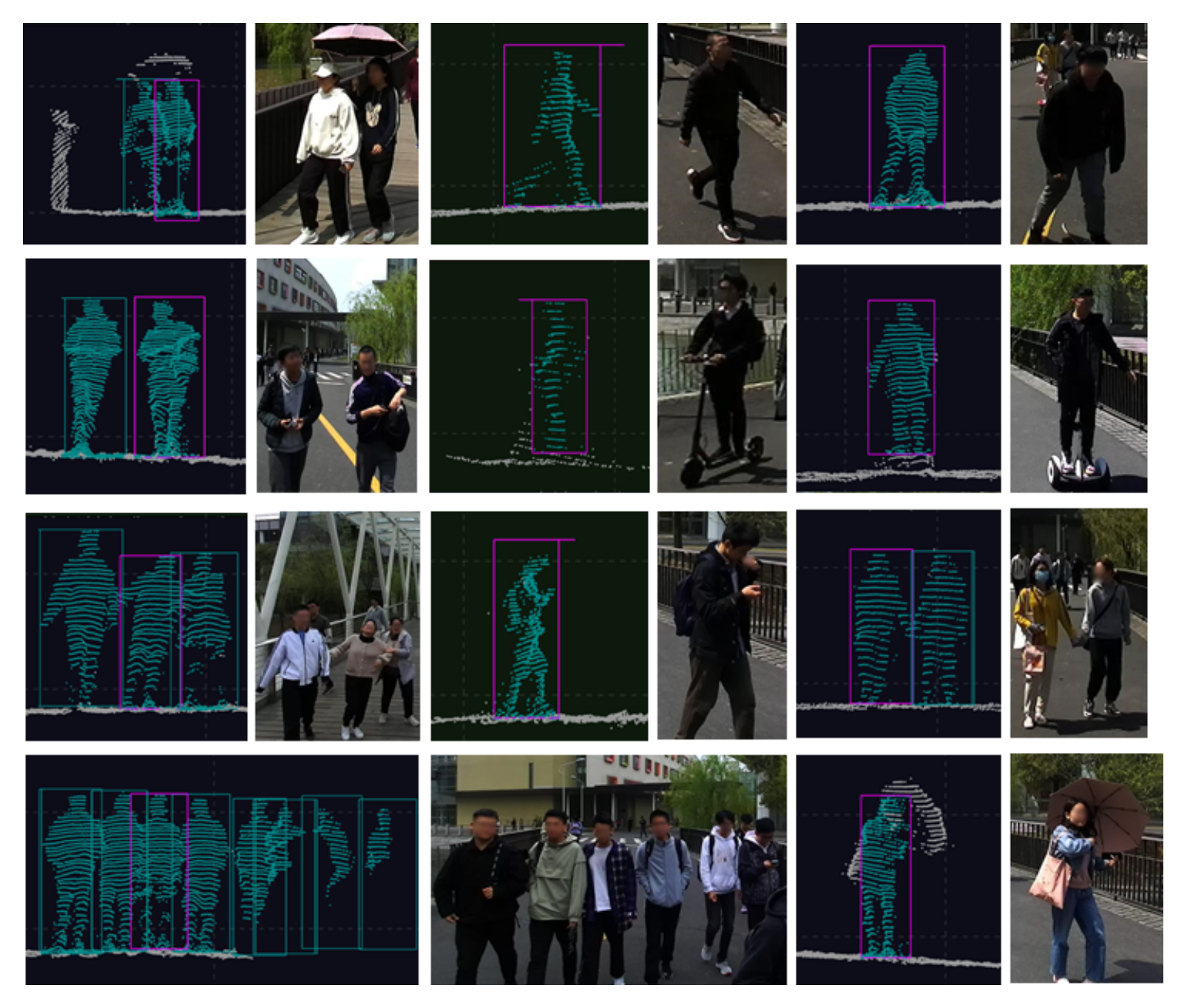}
    \caption{Diverse human poses in semi-structured scenes illustrate the  diversity of pedestrian postures, where pedestrians interact with both dynamic and static elements.}
    \label{fig: poses}
\end{figure}

To address these limitations, PFSD’s semi-structured scenes like \Cref{fig:detection} combine structured elements (e.g., roads, buildings) with dynamic objects (e.g., pedestrians, vehicles). Key challenges include high human participation, dynamic changes, and unpredictability, especially in city streets and public spaces. Dense pedestrian distribution increases occlusion complexity shown in \Cref{fig:occlusion} and interaction challenges shown in \Cref{fig: poses}, requiring models to differentiate adjacent or crossing targets and avoid misclassification.

\begin{figure*}[htbp]
    \centering
    \includegraphics[width=18cm, height=7.5cm]{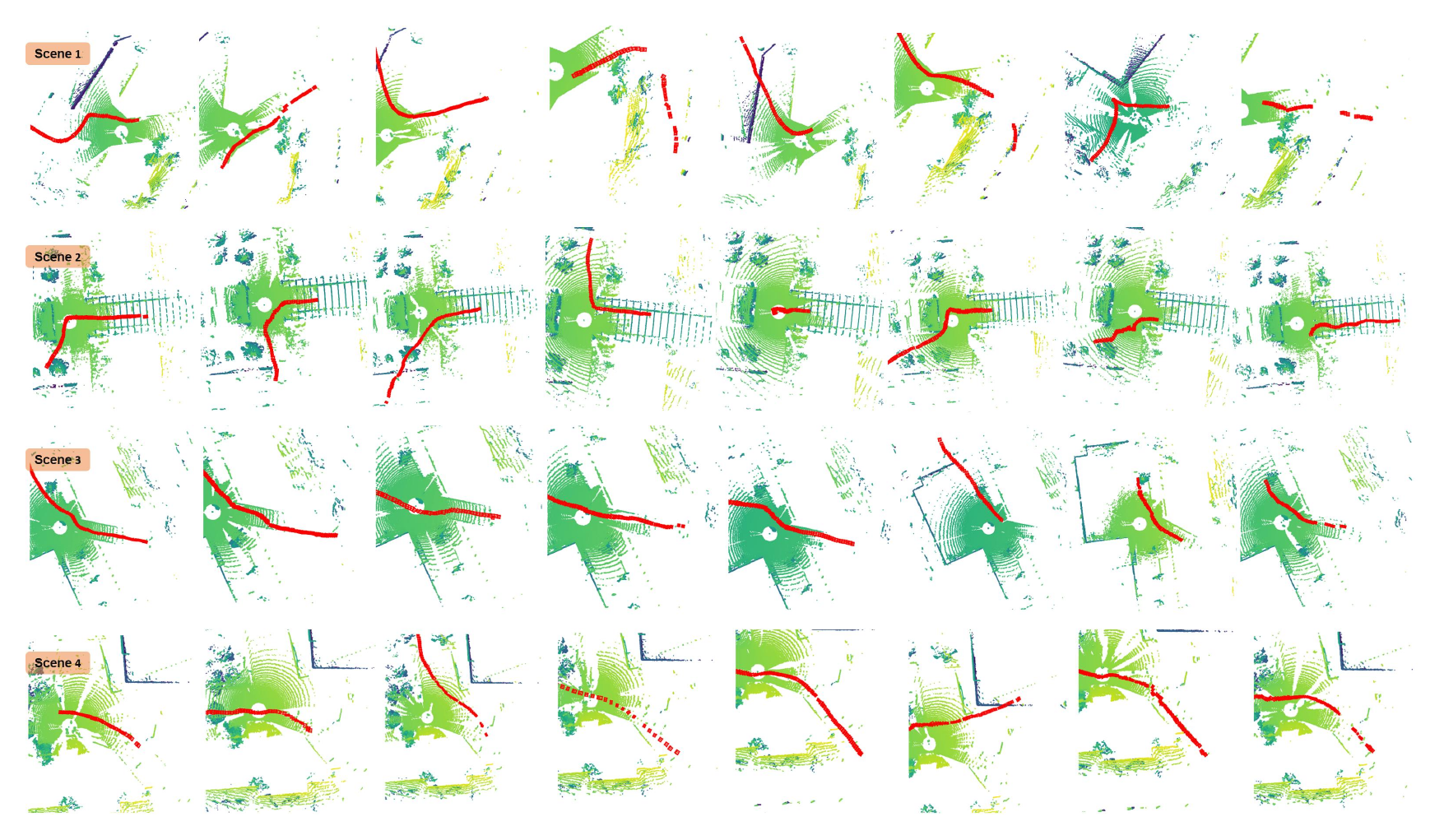}
    \caption{Instance-level pedestrian tracking in semi-structured scenes shows the action trajectory of instances in different scenes, where instances may have certain degrees of occlusion in some point clouds.}
    \label{fig:tracking}
\end{figure*}

In the semi-structured scenes of PFSD, increased pedestrian density is accompanied by greater background complexity and dynamics. Models must handle not only fixed-background targets but also dynamic ones like pedestrians and vehicles. Compared to datasets like KITTI~\cite{kitti}, ApolloScapes~\cite{huang2018apolloscape}, Argoverse~\cite{chang2019argoverse}, A*3D~\cite{A3D}, A2D2~\cite{a2d2}, Cirrus~\cite{cirrus}, and ONCE~\cite{once}, PFSD features significantly higher pedestrian density, as shown in \Cref{tab:datasets}. Pedestrian density in semi-structured scenes is significantly higher, averaging 32 per frame compared to fewer than 5 in most datasets. Moreover, within a 2m range, PFSD achieves a density of 2.6, far exceeding other datasets, and at 5m and 10m ranges, its density is double or more compared to the highest in others. Even H3D, designed for crowded urban scenes~\cite{h3d}, has only half the pedestrian density of PFSD.

\subsection{Instance Tracking in Semi-Structured Environments}
In object tracking tasks, the design of the dataset needs to reflect the dynamic changes and interactions of targets in real-world scenarios. 

Structured scenes feature simple target movements, stable backgrounds, and limited object interactions, as constrained by traffic rules, where Object trajectories are often discrete and distorted. In contrast, unstructured scenes have fewer foreground objects and more irregular backgrounds.

Semi-structured scenes effectively reflect complex relationships between objects, such as crowd behavior, obstacle avoidance, and the clustering effect, which are crucial for object tracking. Semi-structured scenes retain structural elements while introducing dynamic changes, requiring models to exhibit greater flexibility and robustness to handle complex interactions between targets and dynamic backgrounds. For example, as seen in \Cref{fig:tracking}, our dataset features more coherent and longer tracking trajectories which often exhibit high uncertainty and unsmooth. Rule constraints for pedestrians and vehicles are weaker, and multiple targets may become occluded or overlap. In the point cloud data, occlusion is particularly evident, with striped blank areas indicating target occlusion as the last two examples of scene 4 in \Cref{fig:tracking}, which may disrupt tracking continuity. This requires the model to not only track single targets but also handle crowd behaviors and target aggregation, which could help improve the model's performance in multi-object tracking and long-term tracking tasks. 

\begin{figure}[t!]
    \centering
    \includegraphics[width=8.5cm, height=6cm]{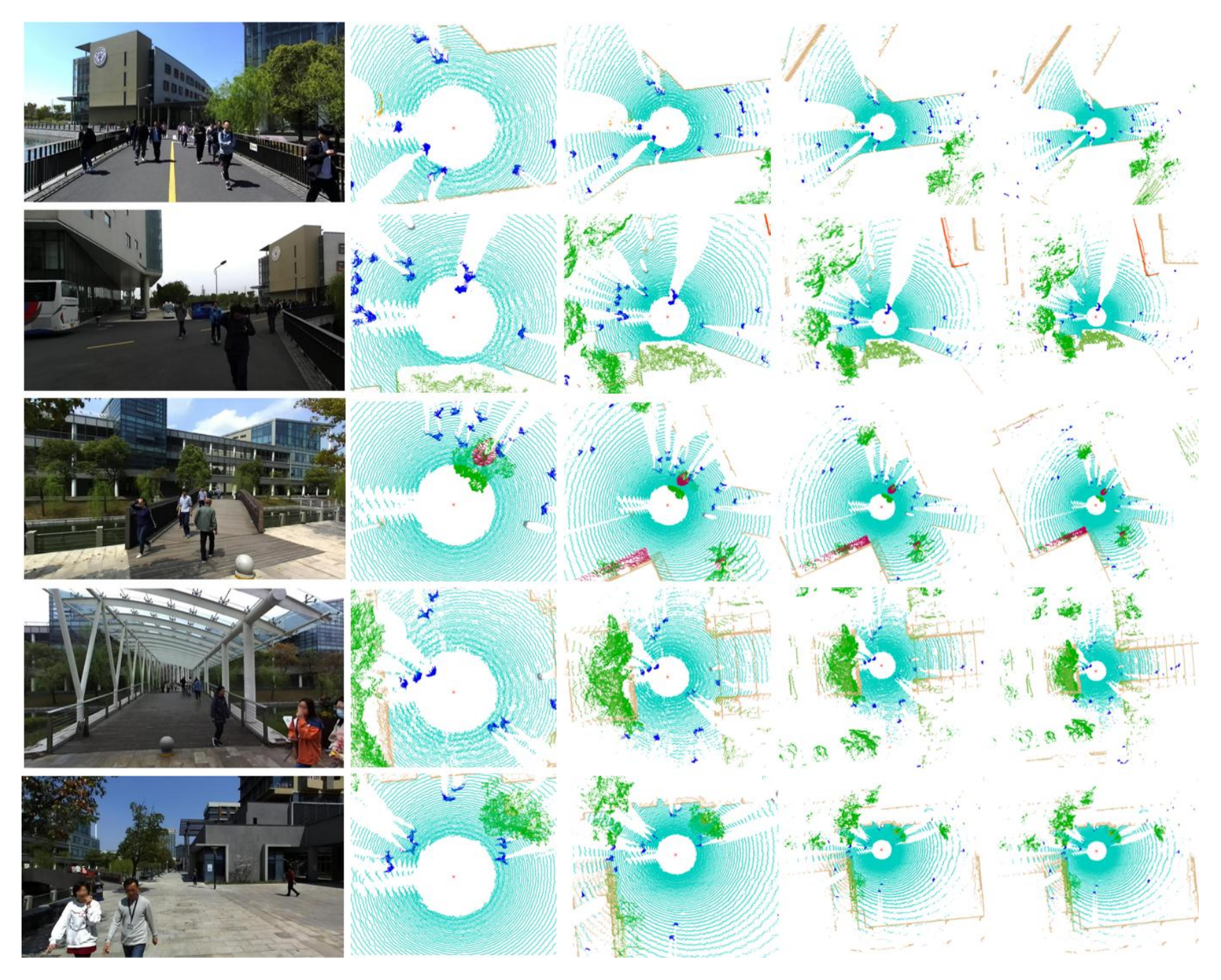}
    \caption{The visualization shows semantic segmentation of various semi-structured scenes, which also indirectly shows the dense pedestrians in our benchmark. Each color represents a different semantic category. }
    \label{fig:segmentation}
\end{figure}

\begin{figure*}[t!]
    \centering
    \includegraphics[width=18cm, height=5.5cm]{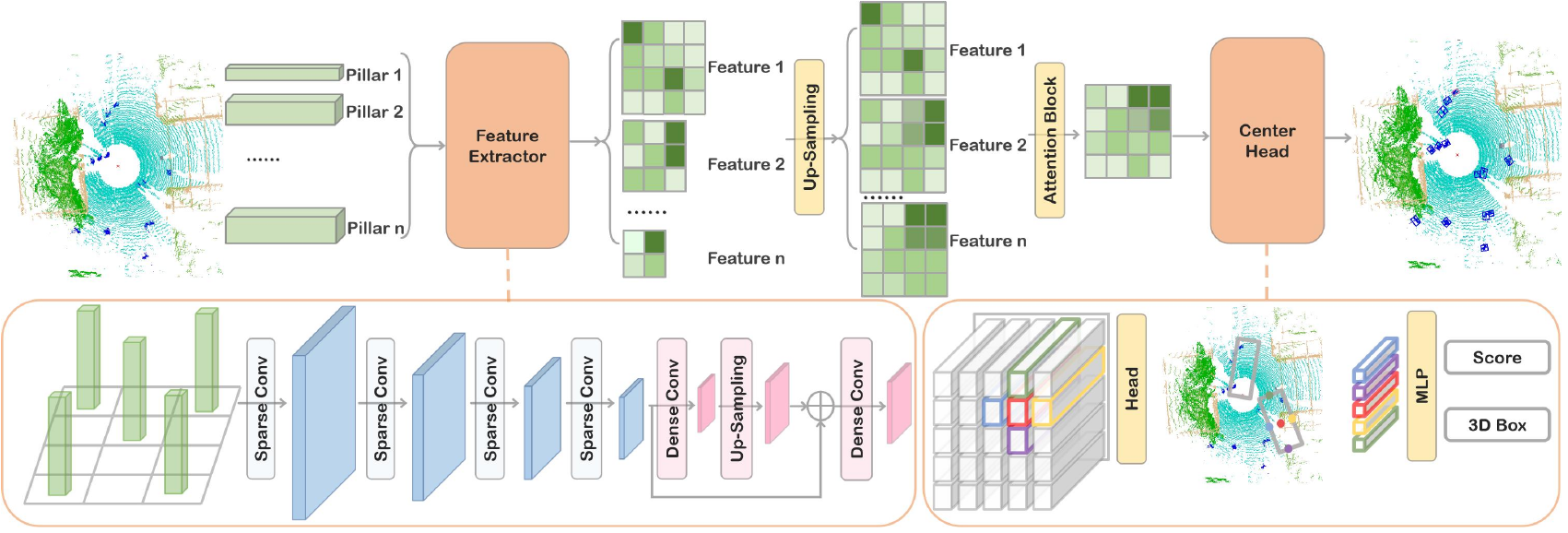}
    \caption{Multi-feature Fusion Pillar-based Network (HMFN) architecture, which integrates multi-scale feature extraction and attention-based fusion techniques to better capture fine-grained details in semi-structured scenes while maintaining computational efficiency.}
    \label{fig:HMFN-arch}
\end{figure*}

\subsection{Segmentation in Semi-Structured Environments}
In object segmentation tasks, the complexity of the dataset scene directly affects the model's ability to learn fine details and handle complex backgrounds. 

Structured datasets typically come from environments with clear boundaries, like urban roads and highways, where object distributions are fixed. Unstructured datasets, such as RELLIS-3D~\cite{rellis}, feature random backgrounds and object distributions, requiring models to find suitable initial values in a vast search space and lacking foreground annotations.

In contrast, PFSD focuses on semi-structured scenes that combine the regularity of structured environments with the variability of unstructured environments, where there are different categories of pedestrians, vehicles, and buildings. In PFSD, shown in \Cref{fig:segmentation}, roadways and streets have a strong sense of integration with surrounding buildings, reducing the independence and prominence of roads found in structured datasets. Pedestrians and vehicles share some roads, and the surrounding urban infrastructure and beautification designs (such as pedestrian bridges, plants, and construction sites) further increase the complexity and diversity of the scenes. Unlike structured datasets that focus on vehicles as the foreground, PFSD places a greater emphasis on pedestrians. PFSD provides more comprehensive training data and richer scene complexity for segmentation tasks.

Besides, by utilizing segmentation results to refine 3D bounding boxes, we enhance the precision of pedestrian localization in high-density scenarios. The segmentation data allows for a more granular understanding of pedestrian distribution, movement patterns, and spatial relationships, ultimately reducing bounding box ambiguity and improving detection robustness. Then, this enhancement contributes to the improvement of pedestrian trajectory tracking, enabling more precise and stable motion prediction. By refining both pedestrian detection and trajectory estimation, our dataset offers a more reliable benchmark for evaluating algorithms, paving the way for more effective and generalizable pedestrian tracking, trajectory forecasting, and behavior analysis in real-world scenarios.  Furthermore, by incorporating segmentation, we significantly enhance the versatility and applicability of the original dataset to allow for the evaluation of segmentation, detection, and tracking tasks within semi-structured dense pedestrian scenes. 

\section{HMFN: Hybrid Multi-Scale Fusion Network}
To tackle the challenges of detection in dense semi-structured scenes with occlusions and scale variations, we propose the Hybrid Multi-Scale Fusion Network (HMFN), as shown in \Cref{fig:HMFN-arch}. Our model uses PillarNet~\cite{shi2022pillarnet} to extract fine-grained features at different scales, fuses them through upsampling and attention mechanisms, and generates a rich feature map for CenterHead, which predicts target centers and sizes directly, effectively handling occlusions and overlaps. Unlike the original PillarNet~\cite{shi2022pillarnet}, which relies on a fixed pillar size, HMFN’s multi-scale fusion enhances feature extraction, addressing occlusions in crowded semi-structured scenes like PFSD. This approach can also extend to segmentation and tracking tasks, where multi-granularity features improve precision in distinguishing overlapping objects.

\subsection{Pillar Encoder}

Referring pillarNet~\cite{shi2022pillarnet}, to encode spatial information from the 3D point cloud in semi-structured scenes, the sparse point cloud $P \in \mathbb{R}^{N \times D}$, where $N$ is the number of points and $D$ is the feature dimension, is divided into $M \times K$ pillars on a Bird’s Eye View plane. Each pillar’s spatial resolution is determined by $(P_m, P_k)$ in the $x$ and $y$ directions.

For each point within a pillar, positional offsets $(\Delta x, \Delta y)$ relative to the pillar center are computed and concatenated with the original features to form $p_i \in \mathbb{R}^{D+2}$, as shown in Equation~(\ref{eq:1}). A linear transformation $W_f \in \mathbb{R}^{(D+2) \times E}$ encodes these features into $f_i \in \mathbb{R}^{E}$, and the final pillar feature $f_{\text{pillar}}$ is obtained by aggregating all point features within the pillar, as shown in Equation~(\ref{eq:3}).

\begin{equation} 
p_i = \left[ \text{features}_i, \Delta x_i, \Delta y_i \right]  \label{eq:1} 
\end{equation}
\begin{equation} 
f_i = W_f p_i,  
f_{\text{pillar}} = \text{Fusion}( { f_1, f_2, \dots, f_n } ) \label{eq:3} 
\end{equation}

\subsection{Pillar Multi-Feature Extraction}
To capture information across multiple spatial scales from semi-structured scenes, we configure pillars of various sizes, denoted as $P_1, P_2, \dots, P_n$. Each scale is processed through a sparse convolutional network that produces deep feature maps, followed by standard convolutions to refine these maps. The resulting sparse feature maps $S_i$ undergo a series of down-sampling operations to capture different spatial resolutions. Standard convolution is then applied to transform the sparse maps into higher-resolution feature maps $C_i$.

Multi-scale features are combined using weighted summation or concatenation to generate the final feature map $F_i$ for each pillar size $P_i$:

\begin{equation} F_i = \text{Fusion}(S_i, C_i) \label{eq:4} \end{equation}

To unify the feature maps across different resolutions, we apply up-sampling using interpolation methods (e.g., bilinear or nearest-neighbor), retaining essential fine details. These up-sampled feature maps are then passed through an attention mechanism, which weights each scale's contribution and produces a fused feature map $F$ as in Equation~(\ref{eq:5}).

\begin{equation} F = \text{AttentionFusion}(F_1, F_2, \dots, F_n) \label{eq:5} \end{equation}

\section{Experiments}
\subsection{Experimental Setup}
We conduct experiments on the PFSD benchmark to validate the robustness of HMFN. PFSD offers over 130,000 pedestrian annotations across diverse semi-structured scenes with varying densities, making it particularly suitable for benchmarking pedestrian detection in complex scenarios. 

\textbf{Dataset Setting.} We conducted every experiment on our PFSD benchmark, dividing it based on different scenarios. Following the proportion used in nuScenes, we split the dataset into training, validation, and test sets with a ratio of 70\%/15\%/15\%. The scenes can be broadly categorized into five types based on the distribution of buildings, like main pathway scenes, courtyard scenes, bridge crossing scenes, covered corridor scenes, and open plaza scenes. At least four of these categories are represented in both the validation and test sets, ensuring a balanced and diverse distribution across various scene types. This strategy minimizes data bias and improves the reliability of performance evaluation. 

\textbf{Implementation Details.} For implementing HMFN, we train the model on an NVIDIA V100 GPU with 32 GB memory. The model is trained for 20 epochs with a batch size of 4, using the Adam optimizer and a one-cycle learning rate policy. The initial learning rate is set to 0.001, with a weight decay of 0.01. Gradient clipping is applied with a norm threshold of 10 to stabilize training. As for other baseline methods, we have followed their original parameter settings on the OpenPCDet framework or provided by authors.

\begin{table}[t!]
\centering
\caption{Performance comparison on PFSD. Bold numbers indicate the best performance for each evaluation metric. The table presents the Average Precision (AP) at different IoU thresholds (0.5, 1.0, 2.0, and 4.0) as well as the mean Average Precision (mAP) across all thresholds.}
\label{tab:overall_comparison}
\setlength{\tabcolsep}{6pt} 
\renewcommand{\arraystretch}{1.3} 
\resizebox{\linewidth}{!}{  
\begin{tabular}{lccccc}
\hline
Method       & AP@0.5 & AP@1.0 & AP@2.0 & AP@4.0 & mAP   \\ \hline
PointPillar~\cite{lang2019pointpillars}  & 21.41  & 33.70  & 55.58  & 74.66  & 45.83 \\
SECOND~\cite{yan2018second}       & 22.75  & 36.53  & 53.21  & 69.08  & 45.40 \\
PillarNet~\cite{shi2022pillarnet}    & 63.25  & 70.57  & 78.20  & 86.13  & 74.52 \\
VoxelNet~\cite{zhou2018voxelnet}     & 69.03  & 75.03  & 81.68  & 87.68  & 78.35 \\
CenterFormer~\cite{CenterFormer} & 71.21  & 76.07  & 82.20  & 88.03  & 79.37 \\
TransFusion-lidar~\cite{transfusion}  & 71.43  & 78.42  & 83.69  & 87.59  & 80.28 \\
VoxelNeXt~\cite{VoxelNeXt}    & 74.35  & 79.78  & 84.33  & 88.01  & 81.61 \\
\textbf{HMFN (Ours)} & \textbf{78.83} & \textbf{82.53} & \textbf{86.99} & \textbf{91.26} & \textbf{84.96} \\ \hline
\end{tabular}
}
\end{table}

\begin{table*}[h!]
\centering
\caption{Ablation studies on HMFN. Pedes-only refers to training with pedestrian annotations exclusively. The table presents several variants including the effect of voxel size, the impact of feature fusion, and pedestrian-specific training.}
\label{tab:ablation_studies}
\resizebox{\textwidth}{!}{%
\tiny
\begin{tabular}{ccccc|ccccc}
\hline
\textbf{Model} & \textbf{Voxel Size} & \textbf{Speed} & \textbf{Order} & \textbf{Pedes-only} & \textbf{AP@0.5} & \textbf{AP@1.0} & \textbf{AP@2.0} & \textbf{AP@4.0} & \textbf{mAP} \\ \hline
PillarNet (Base) & 0.075        & \(\times\)  & -        & \(\times\)  & 60.88       & 68.45       & 76.77       & 85.48       & 72.82      \\
1                & 0.075+0.05   & \(\times\)  & \(\downarrow\) & \(\times\)  & 55.17       & 62.42       & 71.63       & 82.43       & 67.98      \\
2                & 0.075+0.05   & \(\checkmark\) & \(\downarrow\) & \(\times\)  & 43.20       & 51.61       & 62.60       & 77.75       & 58.79      \\
3                & 0.075+0.05   & \(\times\)  & \(\downarrow\) & \(\checkmark\) & 70.87       & 76.00       & 82.08       & 88.69       & 79.40      \\
4                & 0.075+0.10   & \(\times\)  & \(\uparrow\) & \(\checkmark\) & 64.24       & 69.56       & 77.47       & 86.08       & 74.33      \\
5                & 0.075+0.15   & \(\times\)  & \(\uparrow\) & \(\times\)  & 54.86       & 62.74       & 71.83       & 83.40       & 68.21      \\
HMFN (Ours)      & 0.05+0.075   & \(\times\)  & \(\uparrow\) & \(\checkmark\) & \textbf{78.83} & \textbf{82.53} & \textbf{86.99} & \textbf{91.26} & \textbf{84.96} \\ \hline
\end{tabular}%
}
\end{table*}

\subsection{Overall Performance Comparison}
The quantitative comparison of HMFN against several methods on the PFSD benchmark can be found in \Cref{tab:overall_comparison}. Our proposed HMFN achieves significant improvements across all IoU thresholds (AP@0.5, AP@1.0, AP@2.0, AP@4.0) and the overall mean Average Precision (mAP), setting a new benchmark for pedestrian detection performance. Specifically, HMFN achieves an AP@0.5 of 78.83 and a mAP of 84.96, surpassing the previous best model, VoxelNeXt~\cite{VoxelNeXt}, by 4.48 and 3.35 points, respectively. These results demonstrate the effectiveness of the multi-scale feature fusion and attention mechanisms in addressing challenges such as occlusions, crowded semi-structured scenes, and small-scale object detection.

The performance gap between HMFN and other methods can be attributed to its multi-scale features. Compared to traditional pillar-based methods like PillarNet~\cite{shi2022pillarnet}, which lack multi-scale feature extraction, HMFN effectively captures both fine-grained details and contextual information. While voxel-based methods like VoxelNet~\cite{VoxelNeXt} and VoxelNeXt~\cite{VoxelNeXt} demonstrate strong performance, their reliance on fixed-scale feature extraction limits their ability to handle objects of varying sizes and scales and cost more. Transformer-based approaches such as CenterFormer~\cite{CenterFormer} and TransFusion~\cite{transfusion} improve global feature reasoning but often struggle in scenarios requiring detailed local feature capture. HMFN overcomes these limitations by combining multi-scale feature extraction with attention-based fusion, enabling precise detection across varying conditions. 
At the same time, as shown in the \Cref{tab:overall_comparison}, the performance of PointPillar~\cite{lang2019pointpillars} and SECOND~\cite{yan2018second} is suboptimal. This is primarily because both models rely on PointNet~\cite{qi2017pointnet} to extract only local features within pillars or voxels, lacking the capability to model global context effectively. Additionally, there is notable information loss while processing sparse point clouds. Due to hardware limitations, we adjusted the batch size from 4 to 2 during testing for these models, while keeping the number of epochs unchanged.

\subsection{Ablation Studies}
To investigate the effectiveness of various design choices in HMFN, we conduct a series of ablation studies using the PFSD benchmark. Table~\ref{tab:ablation_studies} summarizes the results, focusing on the impacts of voxel size, feature fusion strategies, and pedestrian-specific training on detection performance.

\textbf{Effect of Voxel Size.}  
The choice of voxel size significantly affects the performance of the model, as it determines the granularity of the representation of the features of the point cloud. The baseline PillarNet~\cite{shi2022pillarnet} with a voxel size of 0.075 achieves an mAP of 72.82. Reducing the voxel size to 0.05+0.075 improves the granularity of feature extraction, enabling HMFN to capture fine-grained pedestrian details better, leading to a significant increase in mAP to 84.96. However, improper order (e.g., 0.075+0.05, Model 1) results in reduced performance (mAP of 67.98).

\textbf{Impact of Feature Fusion and Speed Parameters.}  
The introduction of speed parameters in feature fusion can degrade performance in certain settings. For example, Model 2 incorporates speed features but observes a sharp drop in mAP to 58.79 compared to the baseline. This result suggests that naive integration of speed information and inaccurate speed data may introduce noise or inconsistency.

\textbf{Pedestrian-specific Training.}  
When training exclusively on pedestrian annotations, the model's ability to focus on pedestrian-specific features is enhanced, as evidenced by a performance boost from mAP 72.82 (baseline) to 79.40 (Model 3) and further to 84.96 (HMFN) with optimized voxel size. This result highlights the importance of a pedestrian-focused dataset like PFSD, which provides dense and diverse annotations, enabling the model to generalize well across challenging scenarios.

\section{CONCLUSIONS}

 We provide a  Multi-modal Pedestrian-Focus Dataset (PFSD) and a Hybrid Multi-Scale Fusion Network (HMFN). The combination of PFSD’s rich annotations and HMFN’s advanced design establishes a new benchmark for pedestrian detection in complex dense scenarios.

\addtolength{\textheight}{-12cm}   








\end{document}